\documentclass[sigconf]{acmart}

\setcopyright{rightsretained}

\acmConference[PCI'2018]{Pan-hellenic Conference on Informatics}{November 2018}{Athens, Greece}
\acmYear{2018}
\copyrightyear{2018}

\begin{document}
\title{Towards automated neural design: An open source, distributed neural architecture research framework.}

\author{George Kyriakides}
\orcid{0000-0002-5877-0943}
\affiliation{%
  \institution{University of Macedonia}
  \department{Department of Applied Informatics}
  \streetaddress{Egnatia 156}
  \city{Thessaloniki}
  \country{Greece}
  \postcode{54636}
}
\email{ge.kyriakides@uom.edu.gr}

\author{Konstantinos G. Margaritis}
\affiliation{%
  \institution{University of Macedonia}
  \department{Department of Applied Informatics}
  \streetaddress{Egnatia 156}
  \city{Thessaloniki}
  \country{Greece}
  \postcode{54636}
}
\email{kmarg@uom.gr}

\begin{abstract}
NORD (Neural Operations Research \& Development) is an open source distributed deep learning architectural research framework, based on PyTorch, MPI and Horovod. It aims to make research of deep architectures easier for experts of different domains, in order to accelerate the process of finding better architectures, as well as study the best architectures generated for different datasets.
Although currently under heavy development, the framework aims to allow the easy implementation of different design and optimization method families (optimization algorithms, meta-heuristics, reinforcement learning etc.) as well as the fair comparison between them. Furthermore, due to the computational resources required in order to optimize and evaluate network architectures, it leverage the use of distributed computing, while aiming to minimize the researcher's overhead required to implement it. Moreover, it strives to make the creation of architectures more intuitive, by implementing network descriptors, allowing to separately define the architecture's nodes and connections.
In this paper, we present the framework's current state of development, while presenting its basic concepts, providing simple examples as well as their experimental results
\end{abstract}

%
%



\maketitle

\section{Introduction}

Following the recent developments in deep learning, groups of researchers 
experimented with autonomous generation of network architectures \cite{Mik},\cite{stanley2002evolving},\cite{baker2016designing},\cite{zoph2016neural},\cite{negrinho2017deeparchitect}. 
Although the results are promising, the computational resources 
required exceed those readily available to most academic and many industrial researchers. 
Thus, to make autonomous design of neural architectures widely available, 
more efficient methods have to be developed. Although probable, 
it is unlikely that such methods will be spontaneously created. 
It is more probable that new methods will be increasingly more efficient, 
leading to their incremental adoption, in direct proportion to available
computing resources.

There are many points during an automated design process that can greatly impact 
(and increase) the required resources. The evaluation process 
of an architecture usually demands the most resources. Distributing the workload across 
many machines is certainly lucrative. Either by data parallelization/distribution where each node 
gets a slice of the training data or by model parallelization/distribution where each node gets
 a slice of the network's parameters. The design algorithm itself may present inefficiencies, 
 for example some reinforcement learning algorithms are sample-inefficient \cite{openai_2017}.
 Moreover, it may present opportunities for parallelization for example genetic algorithms are considered 
 "embarrassingly" parallel /cite{Alba1999} . The algorithm that distributes
 the workload among many nodes can itself be a bottleneck, if it is not carefully implemented. 
 
 Naturally, in order to design the different components involved in the design process, expertise 
 in different domains and an understanding of the whole process as a system are required. 
 Outside of dedicated research groups, this is usually difficult to communicate and embrace. 
 Furthermore, even if a single researcher, expert in a field relative to a certain component 
 (ex. meta-heuristics) can understand the process as a system, fellow experts 
 in different fields are needed in other to implement it and conduct research. Thus, a framework that
 allows the independent development of design-process components can greatly accelerate advances 
 in the field.
 
 In this paper we present an open source framework which is currently under development and aims 
 to enable researchers of many different fields to contribute and further our understanding of deep neural architectures. 
 First we present the framework's basic concepts. Following, we provide simple examples, as well as discuss possible further directions and research.
 The framework is available at \cite{github}

\section{Basic Concepts}
\begin{figure}
\includegraphics[width=\columnwidth]{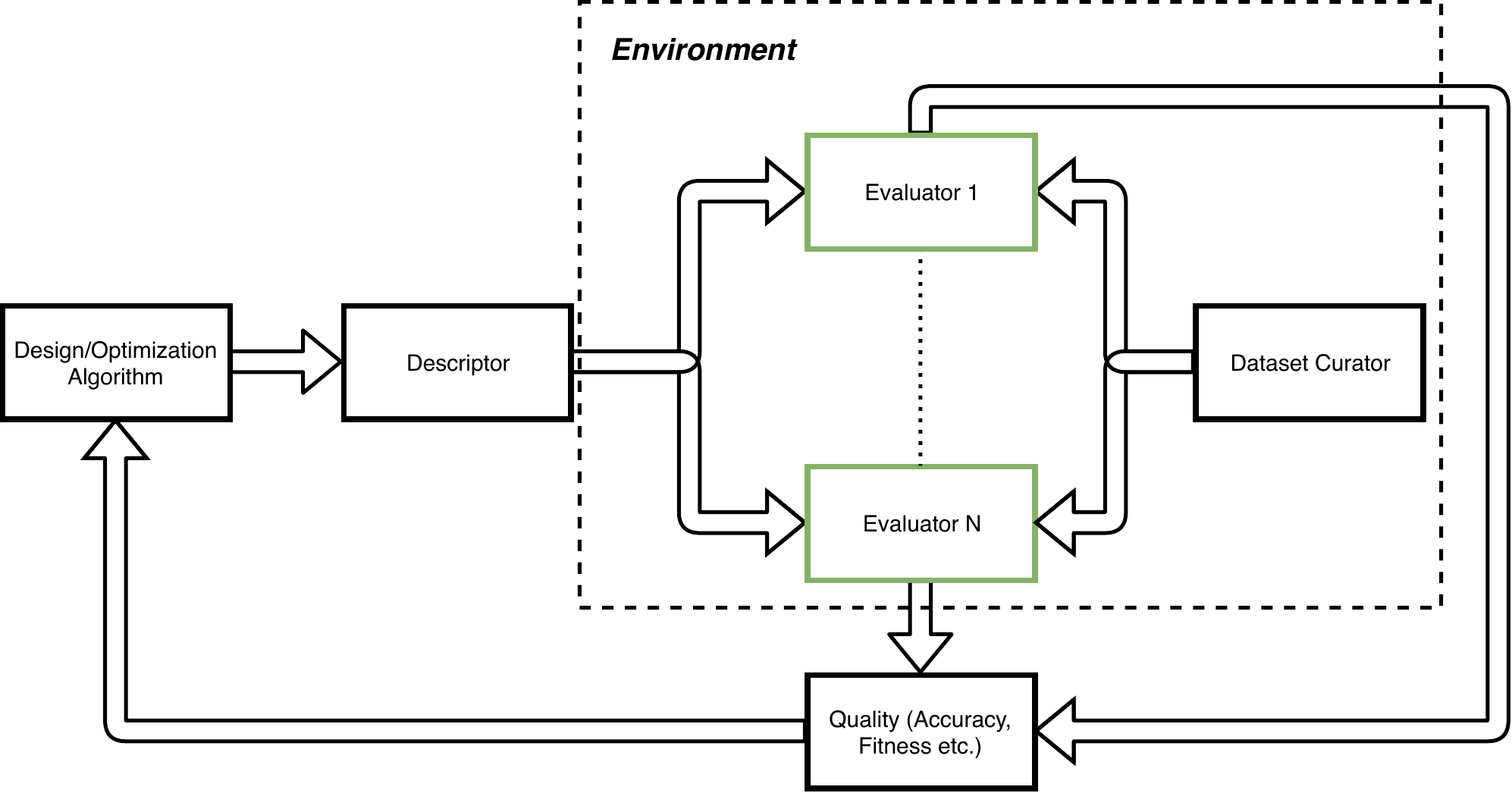}
\caption{Basic workflow for automated architecture design.}
\label{workflow_fig}
\end{figure}

In concordance to many well-established deep learning frameworks, NORD's description of a network 
is based on directed graphs. The framework's main concept is to provide an easy to understand 
programming model in order to efficiently define, optimize and evaluate a graph (i.e. neural network), as well as tools
in order to analyze the generated architectures. In principle, as with any optimization method, the basic workflow 
consists of constructing a solution (network), evaluating it (training and testing) and using the evaluation 
feedback in order to generate another solution, until a certain criterion is met. 
In Figure \ref{workflow_fig} the workflow process is depicted using NORD's basic components.
 
\subsection{Descriptors}
\begin{figure}
\includegraphics[width=\columnwidth]{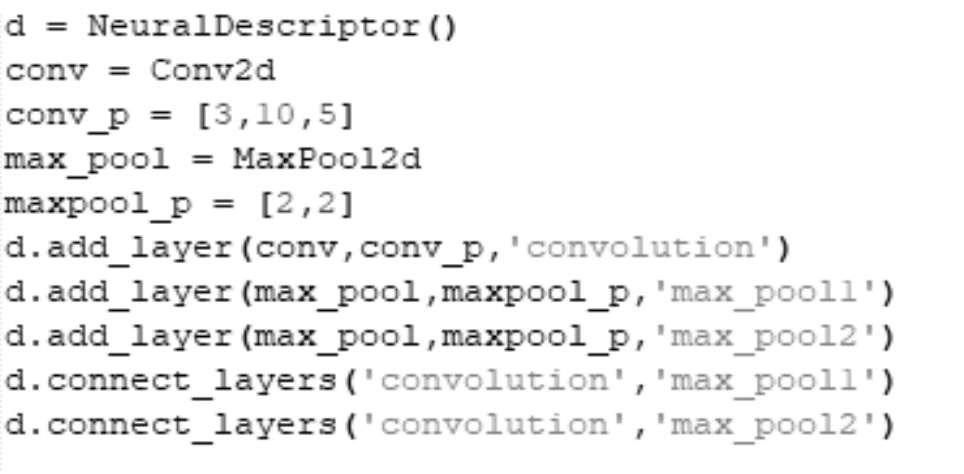}
\caption{Example of a descriptor.}
\label{descriptor_fig}
\end{figure}

In order to make the expression of a network more friendly to automated generation, we implemented descriptors, which allow the independent 
declaration of network nodes and connections within the graph. The first step to define a network using
 descriptors is to add the required layer types, along with their parameters and names. The second step is
 to define connections between the layers, using their names. This aims to make the programmatical 
 and thus, autonomous, creation of complex neural architectures easier and more intuitive. In Figure \ref{descriptor_fig}
 an example of declaring a simple branching convolutional network is given.
 
 Furthermore, a sequential mode can be used, where each layer is automatically connected to the previously added
 layer, making the use of names optional.

\subsection{Evaluators}
The majority of computations required in order to discover good architectures lies within the evaluation 
process. An evaluator is an object that undertakes the process of assessing the quality of a descriptor's architecture. 
An evaluator can have certain properties and operate 
under certain assumptions. For example, it may be executed in a distributed setting or perform certain 
operations in order to reduce the computational cost of evaluating a network. This enables the fair 
comparison between networks and design methods, as it is possible to control and replicate the way that
networks are defined and evaluated.

\subsection{Environments}
An environment defines the distributed setting under which the current evaluator will operate. Using MPI as a reference,
 an Environment tends to the communications initialization, the definition of local and global ranks and 
 finally ensures that the collective communications are terminated correctly. Furthermore, it provides 
 functions to orderly print and log results.

\subsection{Data Curators}
Data curators are responsible for loading evaluation datasets and correctly distributing them amongst the
 workers. Curators provide the required data samplers to evaluators as well as the classes contained 
 in the dataset.

\section{Implementation}
Our current implementation is based on a few well-established libraries, ensuring the correct 
and optimized execution of our framework. We use PyTorch \cite{paszke2017automatic} for the implementation and 
training/evaluation of individual neural networks. In order to distribute the workload of training 
a specific network between many workers, we use Horovod \cite{Sergeev2018}. Finally,to orchestrate the 
collective communication of non-Tensor objects and information we use mpi4py, a fairly robust MPI library for python.

\section{Examples}
In this section we present some simple examples that we implemented to demonstrate the framework's basic concepts, 
as well as to compare their results. The examples aim to design a fully convolutional 
2-layer network and evaluate it on the CIFAR10 dataset \cite{Krizhevsky2009}], using a genetic algorithm. 
The networks generated consist of two convolutional and one classification layer. 
For the first two examples, ESA's PyGMO\cite{Biscani2018} implementation of genetic algorithms was used, while
the third example uses a custom implementation.

\subsection{Local execution}
Our first and simplest example entails the use of a genetic algorithm in order to evolve a small population 
of 2-layer convolutional networks on a single machine. We assume square filters, with sizes of 1 to 50 
pixels, as well as 1 to 50 filters per layer. The genetic algorithm must thus evolve choromosomes with 
a length of 4, representing the size and number of filters for each layer. The algorithm is 
initialized with a crossover rate of 0.9 , mutation rate of 0.2 and uniform sampling from the
possible value range. The population consists of 10 individuals and is evloved for 10 generations.

\begin{figure}
\includegraphics[width=\columnwidth]{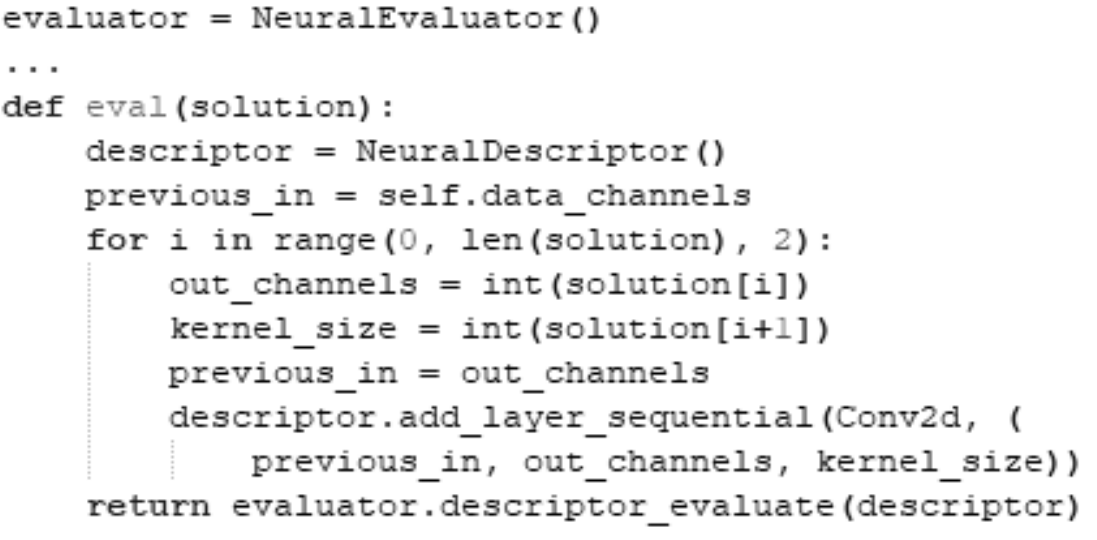}
\caption{Evaluation of a solution.}
\label{local_fig}
\end{figure}

To evaluate each individual, we create a descriptor and add two Conv2d layers, with parameters the first 
and second pair of values contained in the chromosome, using the add\_layer\_sequential function. We then 
pass the descriptor to the evaluator, calling the descriptor\_evaluate function. In Figure \ref{local_fig}
the evaluation code is depicted.

Executed on a single machine with a GT730 graphics card, each generation needed almost 18 minutes to complete.

\subsection{Local execution with distributed evaluation}
This is a typical use case for NORD. The implementation is the same as the above experiment, 
substituting only the evaluator type with the distributed version and executing the code within 
an Environment context manager. By utilizing 9 worker machines, identical to that of the first 
experiment, the average time required for each generation was 6 minutes. Due to the samll size of
our networks, it is probable that more workers were employed than needed. This experiment's
purpose was not to exhibit an almost perfect speedup, but rather to explore the stability and 
ease of implementation for our framework.

\subsection{Distributed execution with local evaluation}
In this example, we implemented a distributed genetic algorithm with local network evaluation. 
Each individual is evaluated on a different machine locally, in contrast to the previous example
where each individual is evaluated by distributing the workload (i.e. dataset) amongst many machines.
In order to achieve this, we implemented a custom genetic algorithm, using MPI for the collective
communication between the workers. The algorithm's parameters remain the same with the previous two
examples.

Using the same execution environment with the previous example, this experiment needed 7 minutes and
40 seconds for each generation on average, being slower than experiment 2. This is due to the fact
that every generation ends when all networks have been evaluated, and thus making big networks 
bottleneck the process.

\subsection{Experimental results}
\begin{figure}
\includegraphics[width=\columnwidth]{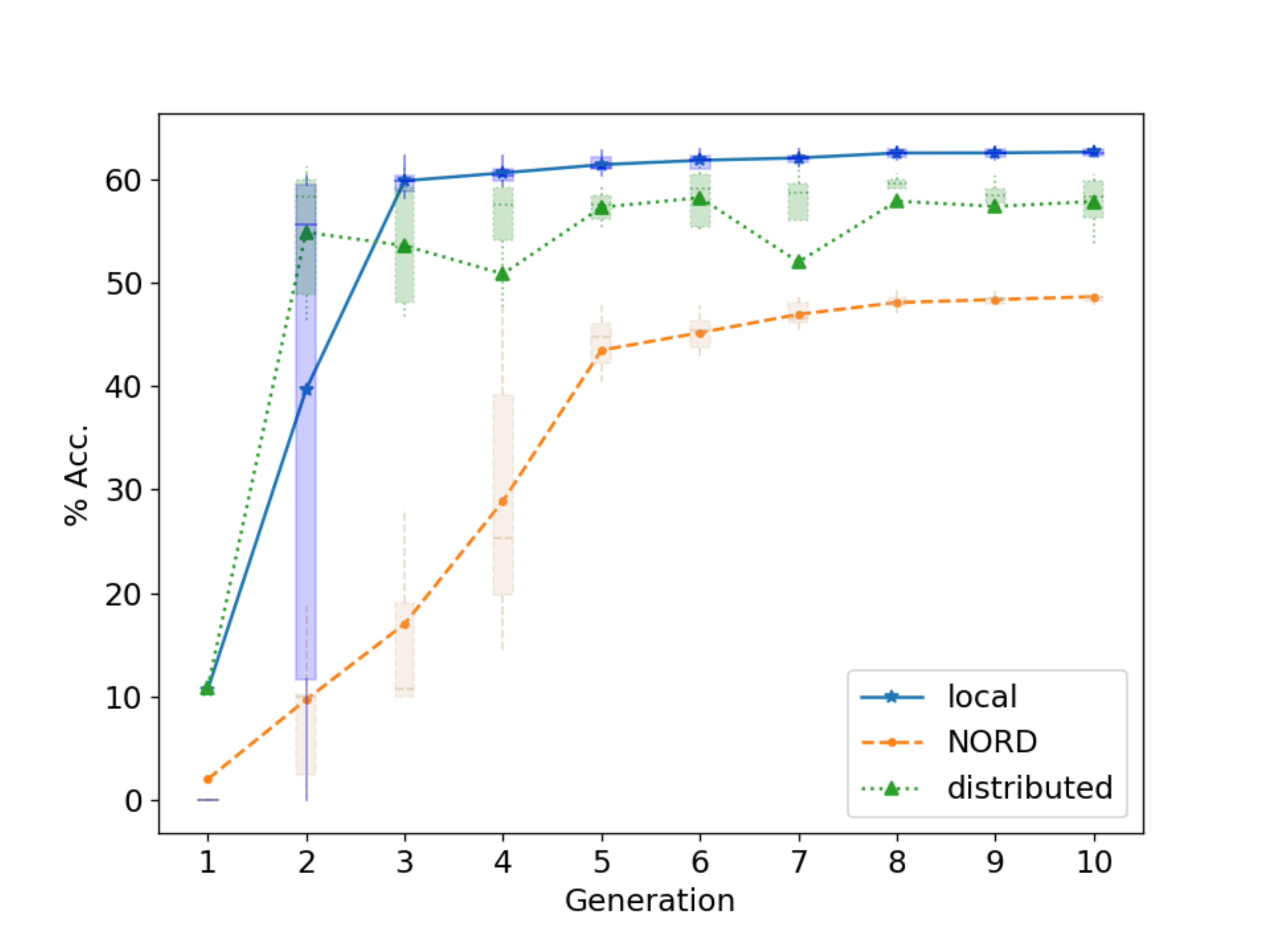}
\caption{Accuracy of solutions for each generation.}
\label{acc_fig}
\end{figure}

In this section we present and compare the results of our framework's current implemented examples.
First, we compare the solutions generated by each method, focusing on the accuracy achieved on the test set.
Figure \ref{acc_fig} depicts the mean accuracy (line plots), as well as the distribution (box plots) of accuracies 
for each generation.
As it is evident, the first example produced the best solutions by running on a single machine. The second 
example, although producing stable solutions did not produce as high quality solutions as the first.
Finally, the third example, although performing worse than the first and better than the second, required
more generations in order to produce stable solutions, as the mean is below the third quantile of the 
solutions' distributions for many generations.

\begin{figure}
\includegraphics[width=\columnwidth]{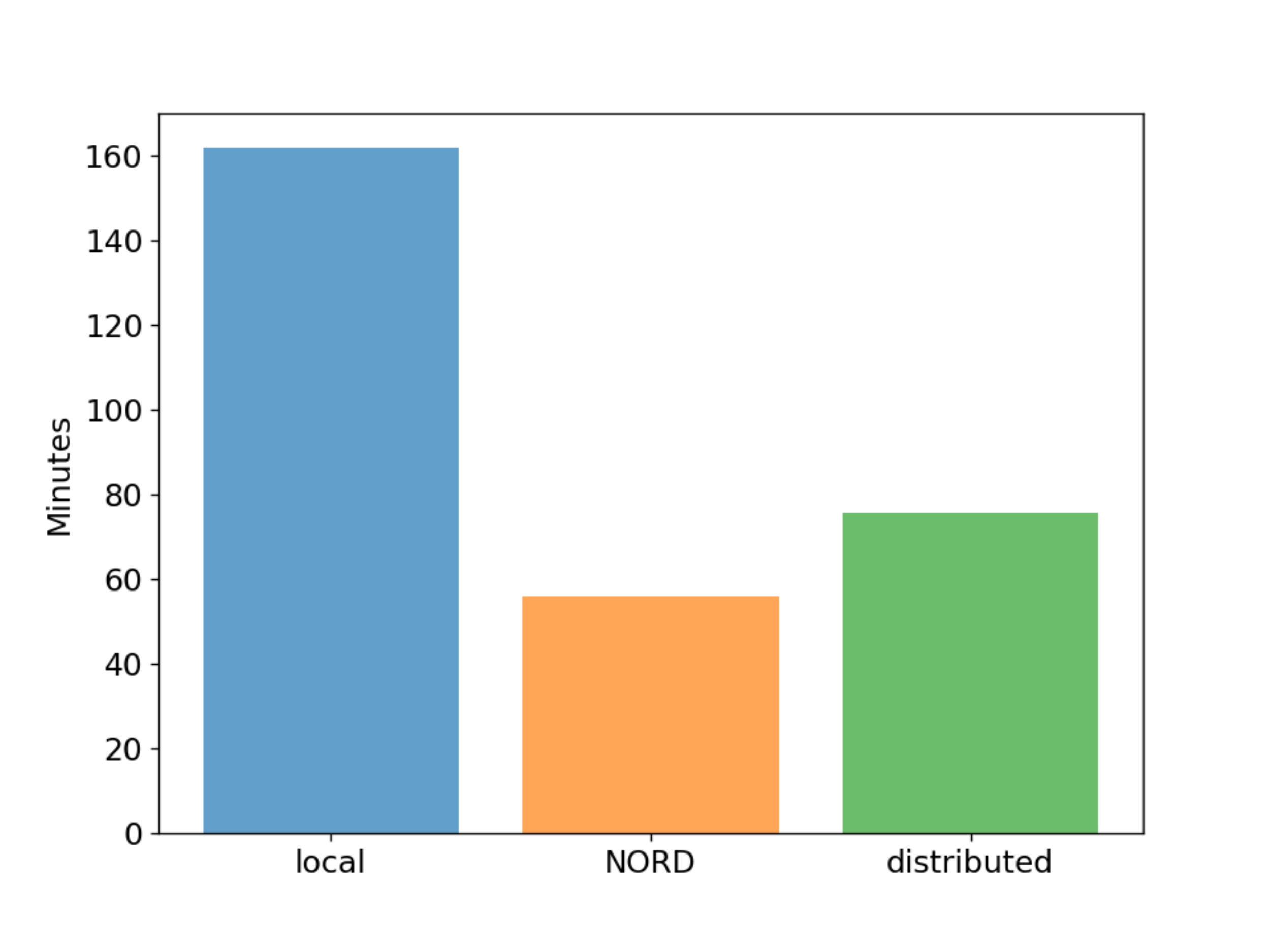}
\caption{Time required for each experiment.}
\label{time_comparison}
\end{figure}

Although the local execution seems to greatly outperform the other two, it can be attributed to the fact 
that the experiments were not executed for many generations due to their requirements in computational resources.
Furthermore, as it is seen in Figure \ref{time_comparison} the local version of our experiments (162 min.) needs almost twice
as much time to complete as the third version (76 min.) and almost three times as much as the 'NORD' version with distributed
evaluation (55 min.).

The distributed version of the experiment, although incurs less communication overhead, requires more time to complete
due to the fact that it is a synchronous algorithm. For each generation, each worker must wait all the others to complete 
before computing the next generation's solution.

\begin{figure}
\includegraphics[width=\columnwidth]{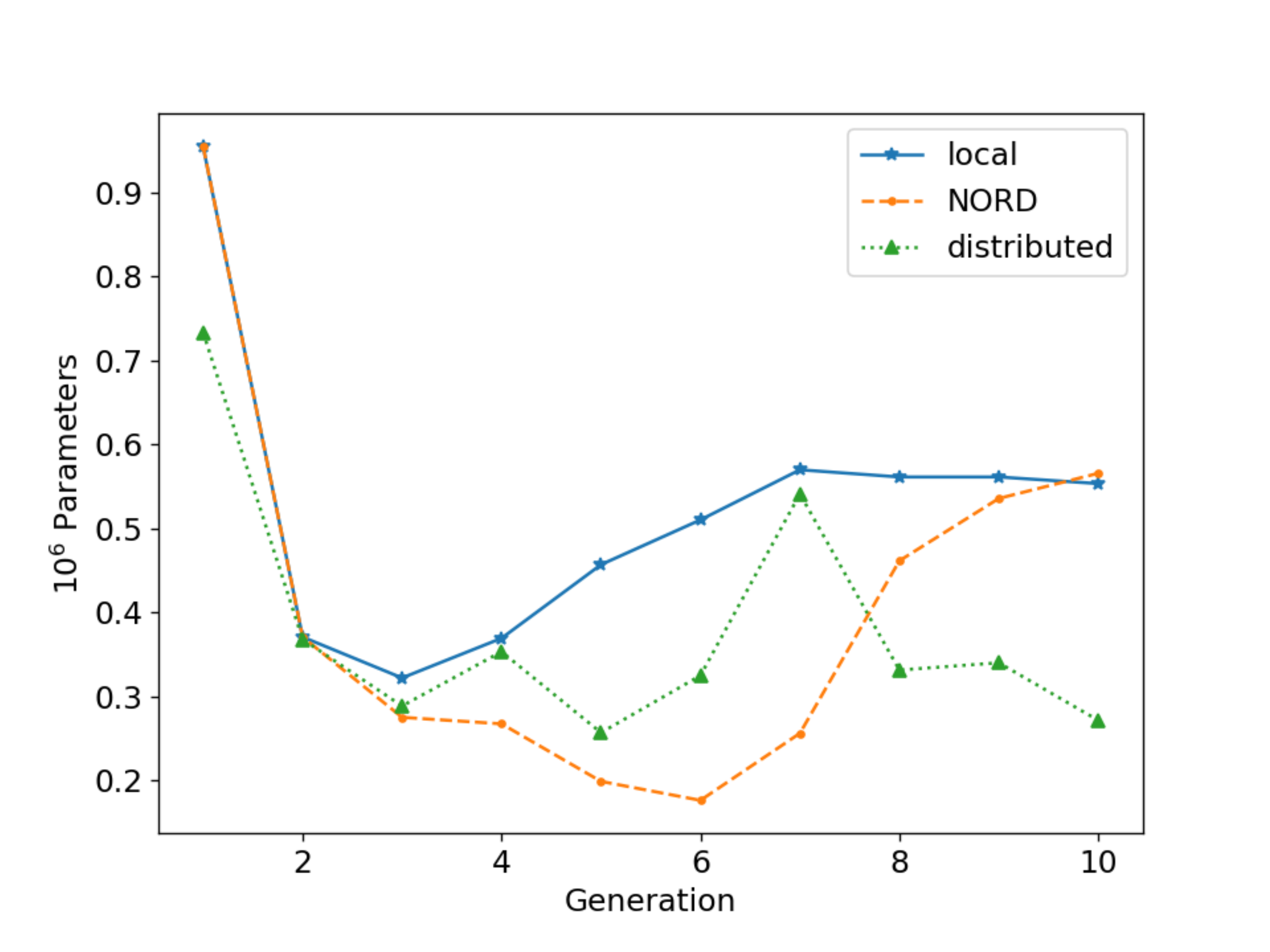}
\caption{Average number of rainable parameters for each generation.}
\label{mean_param_comparison}
\end{figure}

In order to compare the solutions themeselves, we calculate the mean trainable parameters in each generation
for each method. In Figure \ref{mean_param_comparison}, it is evident that the local and 'NORD' versions of 
our experiments, although starting with the same number of parameters, diverge for the majority of the 
generations, only to finally arrive at the same number of average trainable parameters. The distributed version
once again exhibits an unstable behaviour. We should note that the local and 'NORD' versions are
executed using the same PyGMO implementation, while the distributed version uses a custom implementation and thus
certain differences exist.

\section{Limitations and future work}
The framework is currently under heavy development, and many features will be added and refined. Community
feedback and suggestions are welcomed and greatly appreciated. The examples demonstrated here are not meant
to be cutting-edge solutions, but merely working prototypes.
We aim to enrich the framework with tools to provide a fair and thorough implementation, evaluation and
comparison of neural architecture design methods. Furthermore, 

\section {Conclusions}
In this paper we presented a distributed deep learning architectural research framework, based on well-known
and established libraries. We provided some simple examples, by building 2-layer fully convolutional networks for
the CIFAR10 dataset. By implementing a local, distributed and local-execution-distributed-evaluation variant,
we exhibited the basic use of our framework in its current state. By uncoupling the design, evaluation and analysis stages
of the design process, we aim to enable more researchers to contribute to the expansion of our knowledge 
and understadning of deep neural networks.

\bibliographystyle{ACM-Reference-Format}
\bibliography{Arxiv}

\end{document}